\documentclass[runningheads]{llncs}
\usepackage{graphicx}
\usepackage{xcolor}
\usepackage{amsmath}
\usepackage{datetime2}
\usepackage{multirow}
\usepackage{enumitem}
\usepackage{todonotes}
\usepackage{verbatim}
\usepackage{amsmath}

\begin{document}
\title{Machine learning for violence risk assessment using Dutch clinical notes}
\titlerunning{Machine learning for violence risk assessment using Dutch clinical notes}
\author{Pablo Mosteiro\inst{1}\thanks{Corresponding author: p.mosteiro@uu.nl; +31~30~253~9251; Department of Information and Computing Sciences, Buys Ballot Gebouw, Princetonplein 5, 3584 CC Utrecht, the Netherlands} \and
Emil Rijcken\inst{2,1} \and
Kalliopi Zervanou\inst{2} \and
Uzay Kaymak\inst{2} \and
Floortje Scheepers\inst{3} \and
Marco Spruit\inst{1,4,5}}
\authorrunning{P. Mosteiro et al.}
\institute{Utrecht University, Utrecht, the Netherlands \and
Eindhoven University of Technology, Eindhoven, the Netherlands \and
University Medical Center Utrecht, Utrecht, the Netherlands \and
Leiden University Medical Center, Leiden, the Netherlands \and
Leiden Institute of Advanced Computer Science, Leiden, the Netherlands
}
\maketitle
\begin{abstract}
  Violence risk assessment in psychiatric institutions enables interventions to avoid violence incidents. Clinical notes written by practitioners and available in electronic health records are valuable resources capturing unique information, but are seldom used to their full potential. We explore conventional and deep machine learning methods to assess violence risk in psychiatric patients using practitioner notes.
The performance of our best models is comparable to the currently used questionnaire-based method, with an area under the Receiver Operating Characteristic curve of approximately 0.8. We find that the deep-learning model BERTje performs worse than conventional machine learning methods.
  We also evaluate our data and our classifiers to understand the performance of our models better.
  This is particularly important for the applicability of evaluated classifiers to new data, and is also of great interest to practitioners, due to the increased availability of new data in electronic format.

\end{abstract}

\section{Introduction}
Two-thirds of mental health professionals working in Dutch clinical psychiatry institutions report having been a victim of at least one incident of physical violence in their careers~\cite{vanleeuwen}. These incidents can have a strong psychological effect on practitioners~\cite{inoue}, as well as economic consequences for health institutions~\cite{nijman}. Multiple Violence Risk Assessment~(VRA) approaches have been proposed to predict and avoid violence incidents, with some adoption in practice~\cite{singh}. Traditionally, a common approach is the Br{\o}set Violence Checklist (BVC)~\cite{bvc}, a questionnaire used by nurses and psychiatrists to evaluate the likelihood for a patient to become involved in a violence incident. Filling out the form is a time-consuming and highly subjective process.

Machine learning methods might help improve this process by saving time and making predictions more accurate. Electronic Health Records (EHR) are a rich source of information containing both structured fields as well as written notes.
EHR notes coupled with violence incident reports can be used to train machine learning models to classify notes as describing potentially violent patients.
Indeed, machine learning approaches trained on English-language psychiatric notes have shown promising results with values of the Receiver Operating Characteristic Area Under the Curve (AUC) of 0.85 and higher~\cite{chen2018genetic,colling2020predicting,perlis2012using,gorrell2016identifying,MOON201846}.

Nevertheless, violence prediction based on Dutch written notes appears to be a challenging endeavour as no efforts have shown satisfying results, with the AUC stagnating below 0.8~\cite{vincent5,vincent6,his2020}. In those papers, various machine learning methods were applied, including bag-of-words, document embeddings and topic modeling to generate numerical representations of texts; and support vector machines and random forests for classification.
In this work, we explore a new approach to VRA of Dutch clinical notes using the deep-learning language model BERT~\cite{bert}, which has proven to deliver very good results in multiple languages and domains~\cite{quijano2021grid,cui2020revisiting}.
To the best of our knowledge, BERT has not been used for VRA in Dutch before.

The paper is organized as follows. Section~\ref{sec:related_work} is a review of the related work in text analysis for VRA. Sections~\ref{sec:dataset} and~\ref{sec:methodology} describe our dataset and methodology, respectively.
Afterwards, we present our results in Section~\ref{sec:results}, and we discuss our findings in Section~\ref{sec:discussion}. Finally, we present some conclusions in Section~\ref{sec:conclusions}.

\section{Text analysis for Violence Risk Assessment}
\label{sec:related_work}

The analysis of free text in Electronic Health Records (EHR) combined with structured data using machine learning approaches is gaining interest as anonymized EHR data become available for research~\cite{vaci2020natural,colling2020predicting,senior2020identifying}.
However, the analysis of clinical free-text data presents numerous challenges due to (i) \emph{highly imbalanced data} with respect to the class of interest~\cite{Rijoetal2015}; (ii) \emph{lack of publicly available datasets}, limiting research on private institutional data~\cite{Wangetal2018}; and (iii) relatively \emph{small data sizes} compared to the amounts of data currently used in text processing research.  

In the psychiatric domain, structured data such as symptom codes and medication history have been used to predict admissions~\cite{Friedmanetal1983,Lyonsetal1997}.
In combination with structured EHR variables, free text has been used in suicide prediction~\cite{Cooketal2016}; depression diagnosis~\cite{Huangetal2014}; and harm risk in an inpatient forensic psychiatry setting~\cite{violence2018risk}.

To our knowledge, there are few research approaches focusing on predicting violence in mental healthcare based on Dutch-language text from EHRs. Menger et al.~\cite{vincent6} use Dutch clinical text to predict violence incidents from patients in treatment facilities.
In~\cite{his2020} we compared several classical machine learning methods for VRA of EHR notes, including Latent Dirichlet Allocation (LDA) for topic modeling, and we discussed the agreement between some of those classifiers.
In this work we extend on that approach, introducing the BVC as a baseline and employing BERT for document classification.

The general pipeline for violence prediction using Natural Language Processing (NLP) is similar in all approaches. Firstly, the notes are represented in a numerical way. Secondly, the numerical representation is the input to a machine learning algorithm that is trained to perform predictions. 
Van Le et al.~\cite{violence2018risk} use the presence/absence of predefined words as features to feed to several machine learning algorithms. Menger et~al.~\cite{vincent5} experiment with different representations such as bag-of-words, word2vec~\cite{mikolov2013efficient} and paragraph2vec~\cite{paragraph2vec}. Bag-of-words represents documents as vectors with a size equal to the dataset vocabulary length, encoding the presence/absence of a word. The bag-of-words representation is a sparse matrix disregarding word order and meaning. The word2vec algorithm learns word embeddings as dense vectors in high-dimensional space and takes the contexts of words into consideration during training, thus mitigating some of the limitations of the bag-of-words representations. Yet word vectors in word2vec are context-independent, meaning that there is only one vector representation for each word, not taking into account homonyms. The paragraph2vec algorithm is an extension of word2vec that produces a vector representation for an entire paragraph. 

In contrast to word2vec and paragraph2vec, the Bidirectional Encoder Representations from Transformers (BERT) produce a vector representation for each word, taking context into account with higher granularity. By doing so, BERT has improved the state of the art for many NLP tasks~\cite{bert}.

Another method for representing documents is based on topic modeling.
Topic modeling assumes that a document consists of a collection of topics and that each topic consists of a collection of words. After choosing the number of topics to retrieve, the algorithm finds words that contribute most to each topic. A document can then be expressed as a vector, indicating to which extent each topic is represented in that document.
A study in the psychiatry domain~\cite{rumshisky} shows promising results when representing documents using topic models through LDA.
A potential advantage of using topic modeling is that the different topics may help explain how the classification model makes its decisions.
In this paper, we experiment with a model based on LDA topic modeling.

When comparing different models and ranking them, the selection of evaluation metric plays a significant role in the final selection. One of the most commonly used scalars for ranking model performance is the area under the Receiver Operating Characteristic (ROC) curve, typically referred to as Area Under the Curve (AUC)~\cite{Fawcett2006}. 
Different points on the ROC correspond to different operating points of a classifier obtained by thresholding an underlying continuous output that the classifier computes, leading to different false-positive and true-positive rates.
AUC, a scalar measure that takes multiple thresholds into account, is better than accuracy for evaluating the overall classifier performance and discriminating an optimal solution~\cite{huang2005using}. AUC is independent of class priors and it ignores misclassification costs.
In our domain, misclassification costs can be asymmetric, as an unnecessary intervention might be less problematic than an unforeseen violence incident. Hence, it is useful to consider also other performance metrics.

In this paper, we evaluate two alternative metrics: AUPRC and AUK. The Area Under the Precision-Recall Curve (AUPRC) is similar to the AUC, but the axes are precision and recall, which are metrics used more commonly in text classification. AUPRC ignores true negatives, which is desirable when evaluating models on datasets where positives are the minority class and predicting positives correctly is the priority~\cite{SOKOLOVA2009427}.
The Area Under the Kappa curve (AUK)~\cite{kaymak2012auk} is based upon Cohen's Kappa~\cite{cohenskappa}. AUK corrects the accuracy of a model for chance agreements, and it is a non-linear transformation of the difference between a model's AUC value and the AUC of a random model. The main difference between the AUC and AUK is that AUK accounts for class skewness, while AUC does not. Therefore, the AUK has desirable properties when there is considerable class skew.

\section{Data}
  \label{sec:dataset}
  The data used in this study consists of clinical notes written in Dutch by nurses and physicians during visits with patients from the psychiatry ward of the University Medical Center (UMC) Utrecht between 2012-08-01 and 2020-03-01. The 835k notes available are anonymized for patient privacy using DEDUCE~\cite{deduce}.
  The study was reviewed and approved by the UMC ethics committee.

  A patient can be admitted to the psychiatry ward multiple times. Additionally, an admitted patient can spend time in various sub-wards of psychiatry. The time the patient spends in each of the sub-wards is called an \emph{admission period}. In the present study, our data points are admission periods. All notes collected between 28 days before and 1 day after the beginning of the admission period are concatenated and considered a single \emph{period note} for each admission period\footnote{Notes are available from before the beginning of the admission period for patients that have spent time in other wards or sub-wards, or other institutions. For patients transferred within the past 5 days, only notes from 7 or fewer days before the beginning of the admission period are included.}. If a patient is involved in a violence incident between 1 and 28 days after the beginning of the admission period, the outcome is recorded as \emph{violent} (\emph{positive}). Otherwise, it is recorded as \emph{non-violent} (\emph{negative}). Admission periods having period notes with fewer than 100 words are discarded as in previous work~\cite{vincent6,rumshisky}.

  In addition to the notes, we employ structured variables collected by the hospital and related to:
  \begin{itemize}
  \item admission periods (e.g., start date and time);
  \item notes (e.g., date and time of first and last notes in the period);
  \item patient (e.g., gender, age at the beginning of the admission period);
  \item medications (e.g., numbers prescribed and administered);
  \item diagnoses (e.g., presence or absence).
  \end{itemize}
These variables are included to establish whether they correlate with violence incidents.

The resulting dataset consists of 4280 admission periods, corresponding to 2892 unique patients. The dataset is highly imbalanced, as a mere 425 admission periods have a violent outcome.

\section{Methodology}
  \label{sec:methodology}
  In this work, we address the problem of Violence Risk Assessment (VRA) as a document classification task, where EHR document features are combined with additional structured data, as explained in Section~\ref{sec:dataset}.
  Section~\ref{sec:bvc} describes our baseline, the Br{\o}set Violence Checklist.
  In Sections~\ref{sec:ml_analysis} and~\ref{sec:bert_methods} we outline our conventional machine learning and BERT approaches to VRA, respectively. 
We report all our results in three performance metrics: AUPRC, AUC, and AUK as explained in Section~\ref{sec:related_work}. We use AUPRC during development for hyperparameter tuning, while we calculate all three metrics to report the final results.

\subsection{Br{\o}set Violence Checklist}
\label{sec:bvc}
At the University Medical Center (UMC) Utrecht, where our data were collected, VRA is currently done using the Br{\o}set Violence Checklist~(BVC)~\cite{bvc}, a questionnaire that patients answer at the time of admission and approximately every 1-2 weeks during the remainder of the admission. The checklist provides a score from 0 to 6 that estimates the patient's propensity of becoming involved in a violence incident within the next 24 hours.

We use the performance of the BVC as a baseline to compare our machine learning models. The questionnaire was not filled out for all admission periods in our dataset. Therefore, we cross-referenced all the BVC scores in the psychiatry dataset with the violence dataset by patient ID, and took each BVC score as a data point.
The resulting dataset has 11799 datapoints.
The independent variable is the BVC score, and the target variable is the presence or absence of a violence incident within 27 days after the BVC was answered. This mimics the situation in our main analysis, where we consider violence incidents happening between 1 and 28 days after the beginning of the admission period. However, we do not know the time at which the BVC was filled out.
Therefore, we run the analysis twice, once assuming it was filled at 0:00 hours, and another assuming it was filled at the end of the day, at 23:59.

Note that the BVC dataset is larger than the one we used in our machine learning VRA analyses, and its class imbalance is different, with about 1 positive for every 4 negatives (as compared to 1/10 in the VRA dataset). The implications this has on the interpretation of our results is discussed in Section~\ref{sec:results_bvc}.

\subsection{Machine Learning analysis}
\label{sec:ml_analysis}
\subsubsection{Text preparation}
  \label{sec:data_preparation}
  
All notes are pre-processed by applying the following normalization steps:
  \begin{itemize}
  \item converting all period notes to lowercase;
  \item removing special characters (e.g., \"{e} $\rightarrow$ e);
  \item removing non-alphanumeric characters;
  \item tokenizing the texts using the NLTK Dutch word tokenizer~\cite{nltk_book};
  \item removing stopwords using the default NLTK Dutch stopwords list;
  \item stemming using the NLTK Dutch Snowball stemmer;
  \item removing full stops (``.'').
  \end{itemize}
  
  \subsubsection{Text representations}
  \label{sec:text_reps}
  The language used in clinical text is domain-specific, and the notes are rich in technical terms and spelling errors. Pre-trained paragraph embedding models, trained on out-of-domain data, do not necessarily yield useful representations. For this reason, we use our dataset to produce numeric representations for our notes.
We use the entire set of 835k anonymized clinical notes to train both the paragraph embedding and topic models. Only notes with at least 10 words are used in order to increase the likelihood that each note contains valuable information.

\paragraph{Paragraph embeddings} - We use Doc2Vec to convert texts to paragraph embeddings\footnote{Doc2Vec is the name of the Gensim implementation of paragraph2vec.}. The Doc2Vec training parameters are set to the default Gensim~3.8.1 values~\cite{gensim-doc}, except for four parameters: we increase {\tt epochs} from 5 to 20 to improve the probability of convergence; we increase {\tt min\_count} --- the minimum number of times a word has to appear in the corpus to be considered --- from 5 to 20 to avoid including repeated misspellings of words~\cite{vincent6}; we increase {\tt vector\_size} from 100 to 300 to enrich the vectors while keeping the training time acceptable; and we decrease {\tt window} --- the size of the context window --- from 5 to 1 to mitigate the effects of the lack of structure often present in EHR texts.

\paragraph{Topic modeling} -
We use the LdaMallet~\cite{gensim-doc} implementation of LDA to train a topic model. To determine the optimal number of topics, we use the coherence model implemented in Gensim to compute the coherence metric~\cite{syed_coherence}. We find that using 25 topics maximizes coherence. We use default values for the LdaMallet training parameters. Using the trained LDA topic model we compute, for each of the 4280 period notes in our dataset, a 25-dimensional vector of weights, where each dimension represents a topic and the value represents the degree to which this topic is expressed in the note.

  \subsubsection{Classification methods}
  \label{sec:classifiers}
  
  We use two classification methods: support vector machines (SVM)~\cite{svm} and random forest classifiers~\cite{breiman}. For trained random forests, {\tt scikit-learn} outputs a list of the most relevant classification features. This list can help us distinguish which features are most correlated with violence.

\begin{figure}
    \centering
    \includegraphics[width=0.9\textwidth]{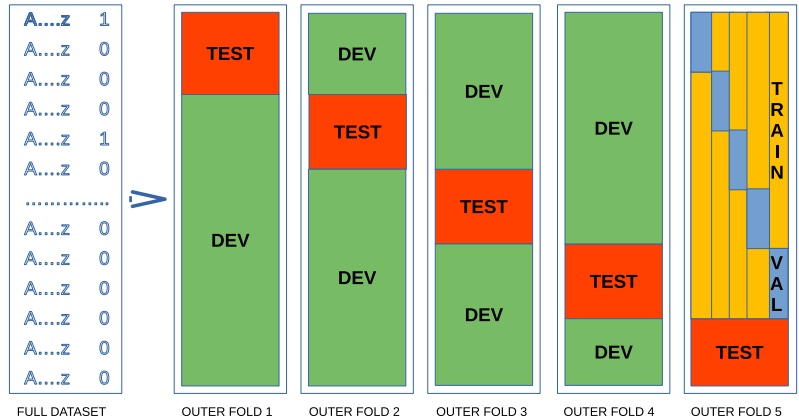}
    \caption{Schematic representation of the data processing pipeline for hyperparameter tuning and statistical uncertainty estimation.}
    \label{fig:fold_diagrams}
\end{figure}

  We use two loops of 5-fold cross-validation for the estimation of uncertainty and hyperparameter tuning, as shown on Figure~\ref{fig:fold_diagrams}. In each iteration of the outer loop, the admission periods corresponding to 20\% of the patients are kept as test data. The admission periods corresponding to the remaining 80\% are the development set.
We then perform 5-fold cross-validation on the development set for hyperparameter tuning; in each iteration, 20\% of patients become the validation set, with the remaining 80\% as the training set.
The combination of hyperparameters that maximizes the average AUPRC on the validation sets in the inner loop is kept. The model is then retrained on the entire development set using the hyperparameters found in the inner loop.
Lastly, the classifier from each iteration of the outer loop is applied to the corresponding test set; we report the average and standard deviation across the outer loop for our three classification metrics.

  For SVM, we employ the support-vector classifier provided by {\tt scikit-learn}, with
  default parameters except for the following: 
  \begin{itemize}
  \item {\tt class\_weight} is set to `balanced', which uses the truth labels to adjust weights inversely proportional to class frequencies, to account for our imbalanced dataset;
  \item {\tt probability} is True to enable probability estimates of performance;
  \item the cost parameter {\tt C} and the kernel coefficient {\tt gamma} are determined by cross-validation.
\end{itemize}
 The ranges of values used are {\tt C} = \{$10^{-1}$, $10^{0}$, $10^{1}$\} and {\tt gamma} = \{$10^{-5}$, $10^{-4}$, $10^{-3}$, $10^{-2}$, $10^{-1}$, $10^{0}$\}, as motivated in a previous study~\cite{vincent6}.

  For the random forest classifier, we use the {\tt scikit-learn} implementation, with default values for all the parameters except for the following: 
  \begin{itemize}
  \item  {\tt n\_estimators} is increased to 500 to prevent overfitting;
  \item {\tt class\_weight} is set to `balanced', which uses the truth labels to adjust weights inversely proportional to class frequencies, to account for our imbalanced dataset;
  \item {\tt min\_samples\_leaf}, {\tt max\_features} and {\tt criterion} are determined by cross-validation.
    \end{itemize}
  The ranges of values used for the parameters determined by cross-validation are reported on Table~\ref{tab:rf_params}. Values for {\tt min\_samples\_leaf} are greater than the default value of 1, to prevent overfitting; we consider the default value of `auto' for {\tt max\_features}, which sets the maximum number of features per split to the square root of the number of features, and two smaller values, again to prevent overfitting; both split criteria available in {\tt scikit-learn} are considered.
\begin{table}
    \centering
\caption{Random forest training parameters. Parameters with multiple values are optimized through cross-validation. Parameters not shown are set to default {\tt scikit-learn} values.}\label{tab:rf_params}
\begin{tabular}{l|l|l}
\hline
Parameter &  Value/s & Method \\
\hline
{\tt min\_samples\_leaf} & \{3, 5, 10\} & Cross-validation \\
{\tt max\_features} & \{5.2, 8.7, `auto'\} & Cross-validation \\
{\tt criterion} & \{`gini', `entropy'\} & Cross-validation \\
\hline
{\tt n\_estimators} & 500 & Fixed \\
{\tt class\_weight} & `balanced' & Fixed \\
\hline
\end{tabular}
\end{table}

  \subsection{BERT analysis}
  \label{sec:bert_methods}
We use the implementation of BERT for sequence classification included in Huggingface Transformers~\cite{transformers}, which includes various choices of pre-trained models. We explored Multilingual BERT~\cite{multilingualbert}, from Google Research, as well as BERTje~\cite{bertje}\footnote{In Huggingface Transformers, BERTje is called {\tt BERT-base-dutch-cased}.}. During an exploratory analysis of the 110kDBRD Dutch book review dataset~\cite{110kDBRD}, where we attempted to predict the sentiment of book reviews, BERTje outperformed Multilingual BERT. Therefore, we decided to continue using BERTje.

  We tokenize all texts using the pre-trained BERTje tokenizer provided by Huggingface Transformers. For classification, a linear layer is added to the BERTje language model. The optimizer used is AdamW, which is recommended by the BERT developers~\cite{bert}. 
  We use a scheduler that increases the learning rate linearly from 0 to the value set in the optimizer during a warm-up period, and then decreases linearly back to 0.
  The warm-up period consists of 10\% of the training set. 
  To choose this number, we ran an exploratory analysis on a dataset of Dutch-language tweets~\cite{joosten}.
  We experimented with warm-up periods of 5\%, 10\% and 20\%, and found the best results for 10\%. 
  Furthermore, we found that a good choice for the learning rate value set in the optimizer was $2\times10^{-5}$. We use the Transformers default dropout probability of 0.1. Finally, we set the {\tt epsilon} parameter in the AdamW optimizer to $10^{-8}$.

  As BERTje can only handle up to 512 tokens in a sequence, we experiment with two shortening strategies to shorten our texts:
  \begin{enumerate}
  \item keeping the last 512 tokens in each period note (\emph{truncate});
  \item summarizing texts using Gensim TextRank (\emph{summarize}).
  \end{enumerate}
  Because tokenization in BERTje splits words into its component morphemes, texts have more tokens than words. Thus, when summarizing texts with Gensim, we choose the summarized length to be shorter than 512 words, so that the tokenized texts have lengths shorter than 512 tokens.

We use two loops of 5-fold cross-validation, as explained in Section~\ref{sec:classifiers} and Figure~\ref{fig:fold_diagrams}, for uncertainty estimation and to tune the batch size; we explore batch sizes of 16, 32, 64, 128 and 256. The optimal number of epochs is chosen as the number of epochs after which the validation loss increases while the training loss decreases; we run all training for a maximum of 4 epochs.

  Finally, because Transformers uses the cross-entropy loss function, which is symmetric to positive and negative samples, we re-balance the training and validation sets before training the model and computing the loss. The performance metrics are computed based on the original (imbalanced) dataset.

  \section{Experimental Results}
  \label{sec:results}
  
  \subsection{Br{\o}set Violence Checklist}
  \label{sec:results_bvc}
\begin{figure}[htb]
  \centering \includegraphics[width=0.9\textwidth]{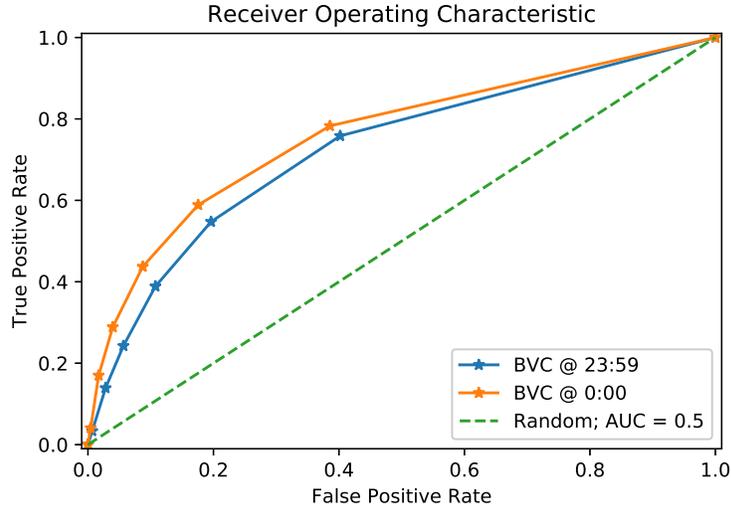}
  \caption{Receiver Operating Characteristic (ROC) of the BVC to predict violence incidents within 27 days after filling out the BVC questionnaire. The orange line assumes the BVC was filled out at the beginning of the day; the blue line assumes the BVC was filled out at the end of the day. The dotted green line corresponds to random guessing.}
  \label{fig:bvc_roc}
\end{figure}

  The predictive power of the BVC for violence risk assessment within 27 days of collecting the questionnaire is shown on Figure~\ref{fig:bvc_roc} and Table~\ref{tab:bvc-results}.
  Because the BVC returns integer predictions from 0 to 6, the ROC, the precision-recall curve, and the kappa curve all have a very small number of points. Thus, the areas under the curves depend heavily on the integration method. We report our central values with integration following the trapezoidal rule. We used the left- and right-hand rules to estimate the uncertainty due to the choice of the integration method.
 \begin{table}
  \centering
\caption{Predictive power of the Br{\o}set Violence Checklist. The uncertainty is given by the choice of the integration method.}\label{tab:bvc-results}
\begin{tabular}{c|c|c|c}
  \hline
  Assumed Filled Hour & AUC & AUPRC & AUK \\
  \hline
  00:00 & 0.761$\pm$0.100 & 0.510$\pm$0.050 & 0.207$\pm$0.090\\
  23:59  & 0.725$\pm$0.107 & 0.391$\pm$0.041 & 0.166$\pm$0.069\\
\hline
\end{tabular}
\end{table}

  Note that amongst the three metrics used, only the AUC can be used for comparison between the BVC and the machine learning models, as the class imbalance of the BVC dataset differs from the distribution of the data based on which the machine learning models are trained (see Section~\ref{sec:discussion} for discussion).
  
  \subsection{Classifier Performance}
  \label{sec:clf_perf}
  
  Table~\ref{tab:results} reports the results of the machine learning models not using BERTje. All configurations gave results consistent with each other, as well as with previous work on a smaller dataset~\cite{vincent6}, and with the BVC.
  \begin{table}[tb]
  \centering
\caption{Classification metrics for various training configurations. The statistical uncertainties were estimated via 5-fold cross-validation.}
\label{tab:results}
\begin{tabular}{l|c|c|c|c}
  \hline
  Text Representation & Classifier & AUC & AUPRC & AUK \\
  \hline
doc2vec & SVM & 0.792$\pm$0.034 & 0.315$\pm$0.075 & 0.136$\pm$0.023 \\
doc2vec & RF & 0.782$\pm$0.030 & 0.287$\pm$0.061 & 0.130$\pm$0.020 \\
doc2vec+struct & RF & 0.777$\pm$0.026 & 0.292$\pm$0.053 & 0.129$\pm$0.017 \\
LDA+struct & RF & 0.785$\pm$0.038 & 0.303$\pm$0.079 & 0.133$\pm$0.025 \\
doc2vec+LDA+struct & RF & 0.792$\pm$0.035 & 0.298$\pm$0.066 & 0.136$\pm$0.022 \\
\hline
\end{tabular}
\end{table}
These metrics show modest performance, and they indicate that further work is needed to extract all the meaningful information contained in the clinical notes.

The results of the BERTje analysis are shown on Table~\ref{tab:bert_results}. The optimal number of epochs was 1 for all configurations. After that, the validation loss increased quickly, while the training loss continued to decrease. This is a sign of overfitting, and we will discuss it further in Section~\ref{sec:discussion}. For comparison, we also trained the models for a second epoch and computed the test metrics again.
\begin{table}
  \centering
\caption{Results of the Violence Risk Assessment analysis using BERTje. The statistical uncertainties were estimated via 5-fold cross-validation.}\label{tab:bert_results}
\begin{tabular}{l|c|c|c|c}
  \hline
  Shortening strategy & Trained epochs & AUC & AUPRC  & AUK\\
\hline
  Summarize & 1 & 0.664$\pm$0.029 & 0.182$\pm$0.049 & 0.071$\pm$0.019 \\
Summarize & 2 & 0.658$\pm$0.033 & 0.184$\pm$0.051 & 0.069$\pm$0.019 \\
Truncate & 1 & 0.667$\pm$0.027 & 0.195$\pm$0.041 & 0.074$\pm$0.017 \\
Truncate & 2 & 0.657$\pm$0.039 & 0.187$\pm$0.045 & 0.070$\pm$0.023 \\
  \hline
\end{tabular}
\end{table}

\section{Discussion}
\label{sec:discussion}

In this work we analyzed how conventional machine learning methods and BERTje performed on Violence Risk Assessment (VRA).
Although, typically, BERT-like models show high performance on similar tasks, we did not find this to be the case with our dataset. BERTje performed significantly worse than the other machine learning models.
None of our models reached performance levels similar to the literature based on English notes. This is a surprising finding, given that we have used more advanced techniques.

In this section we investigate our results further.
Firstly, we compare results from the different performance metrics.
Secondly, we dive deeper into the dataset by studying the type-token ratio.
Next, we examine some problems with our dataset.
Finally, we study the feature importance in the random forest classifiers.
We end this section with a discussion of the limitations of our work.

\subsection{Performance Metrics}
In this work, we have used three different different performance metrics: AUC, AUPRC and AUK.
The AUC is the most widely used metric; since it is invariant under a change in the class imbalance, it gives an idea of the classifier performance regardless of the precise dataset used. However, when choosing among multiple models applied on the same dataset, it is a poor metric because it does not account for the possibility that positive- and negative misclassification might have different costs. The AUPRC and AUK attempt to account for this; the former does so by ignoring true negatives, the latter by accounting for chance agreements between the classifier and the truth labels.

Looking at Tables~\ref{tab:results} and~\ref{tab:bert_results}, we can see that all models within each table performed consistently with each other in all metrics. The relative uncertainty is smallest for AUC.

\begin{figure}
    \centering
    \includegraphics[width=0.9\textwidth]{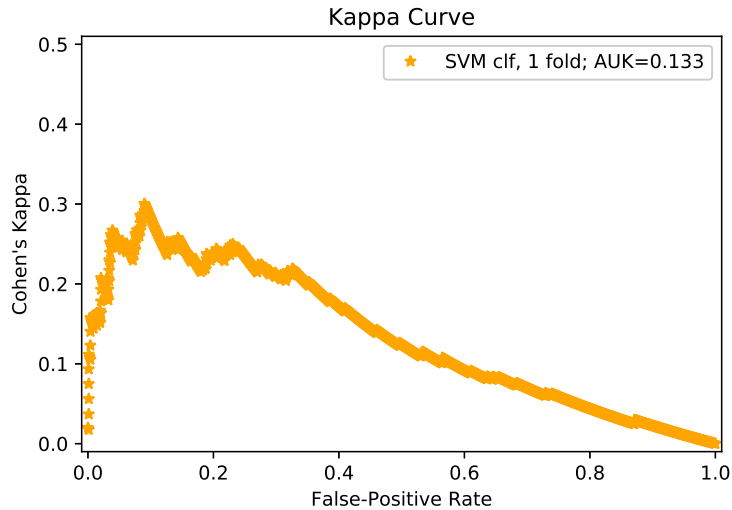}
    \caption{Kappa curve for one of the iterations of the outer cross-validation loop of our SVM classifier.
      The maximum value in this graph can be interpreted as the optimal operating point of the classifier.
      Random guessing results in Cohen's kappa equal to 0 for all false-positive rates.
    }
    \label{fig:kappacurve}
\end{figure}

An advantage of the AUK is that the kappa curve not only provides a performance metric but also suggests an optimal operating point. Figure~\ref{fig:kappacurve} shows the kappa curve for one of the iterations of the SVM classifier. Because the curve is concave down, we can choose the operating threshold that maximizes Cohen's kappa---and thus the performance after accounting for chance agreements--- versus the false-positive rate.
In the figure we can see that there is a wide domain of false-positive rates that give values of Cohen's kappa close to the maximum. In practice, psychiatrists would also inform the choice of what the operating threshold should be, based on the costs of misclassifying positives and negatives.

\subsection{Type-Token Ratio}
\label{sec:ttr}
As mentioned in Section~\ref{sec:clf_perf}, there is evidence that the BERTje models overfit the training data after 2 or more training epochs. To explore why our data might be prone to overfitting, we looked at the type-token ratio (TTR) for this dataset and compared it to that of the 110kDBRD dataset~\cite{110kDBRD},
consisting of book reviews written in Dutch (see Section~\ref{sec:bert_methods}).
The TTR is the ratio of the number of distinct words (types) in a corpus to the total number of words (tokens). The higher the TTR, the more varied the vocabulary.
A varied vocabulary could be associated with overfitting, as the models can exploit individual rare words to differentiate period notes from each other.
We computed the TTR both on the words and on the tokens. Tokenization was done by the pre-trained BERTje tokenizer used in Section~\ref{sec:bert_methods}.

The results are shown on Table~\ref{tab:ttr}. The raw texts from our VRA dataset seem to have a less rich vocabulary than the 110kDBRD dataset; however, after summarization the vocabulary is more varied than before. This could mean that the summarization algorithm picks the most distinctive words and removes very repetitive words. More interestingly, after tokenization both the full texts \emph{and} the summarized texts from the VRA dataset have similar TTRs, while the 110kDBRD data has a much lower TTR. It is expected that tokenization will reduce the TTR, because the tokenizer merges several forms of the same word into its root morphemes. But the effect on the VRA dataset is smaller than on the 110kDBRD dataset. The BERTje tokenizer often splits unknown words into component letters that can have morphemic meaning under some contexts. If our dataset contains more idiosyncratic words that the tokenizer does not know about, it might split them into multiple morphemes that are not really what the word is made up of. This would then lead to classification errors, as the text is interpreted to mean something it does not.
\begin{table}[tb]
  \centering
  \caption{The Type-Token Ratio (TTR) for our dataset, compared to the book-review dataset 110kDBRD. When tokenizing, only the first 512 tokens were considered. The five most common tokens are also reported.
  }\label{tab:ttr}
\begin{tabular}{l|c|c|l}
  \hline
Dataset & TTR (Raw) & TTR (Tokenized) & Most common tokens\\
\hline
VRA (Summarized) & 0.045 & 0.0069 & [PAD], `.', `,', [UNK], \emph{en} \\
VRA (Full) & 0.025 & 0.0076 & `.', [UNK], `,', `-', \emph{en} \\
110kDBRD & 0.052 & 0.0033 & [PAD], `.', \emph{de}, `,', \emph{het}\\
  \hline
\end{tabular}
\end{table}

On Table~\ref{tab:ttr} we see that a large number of raw words in our VRA dataset are interpreted as the [UNK] (unknown) token. Some such words were $<$, $>$, \emph{cq}, \# and +. This could mean that a lot of words are either missed by the tokenizer (incorrectly classified as unknown, with their meaning getting lost), or included unnecessarily ($<$ and $>$ are used at the beginning and end of anonymized institution and person names but carry no meaning). This could contribute to the bad performance of the classifier.

Upon inspecting the cross-validation loss, we found that, after two epochs of training, overfitting was apparent and yet the performance metrics did not significantly worsen. Thus, although overfitting can be a problem with our dataset, it cannot be the reason why our BERTje results are worse than those obtained using simpler machine learning methods.

\subsection{Data Problems}
While exploring our data further, we found that some period notes labeled as {\em non-violent} contained mentions of violence incidents.
To assess the significance of this, we examined all practitioner notes collected between 1 and 28 days after the beginning of every admission period. We curated a list of words related to violence incidents\footnote{An expert psychiatrist curated this list.}. We visually examined some sentences from negative admission periods containing at least one mention of a word from the list and found that other important words had to be added to the list.
The final list of keywords chosen is shown in Table~\ref{tab:unreported_incidents_keywords}.

Simply searching for a keyword can lead to too many false positives, because some words have multiple meanings. This is especially true in Dutch, where certain phrasal verbs have completely different meanings than their root verbs (e.g., \emph{slaan} (hit) vs \emph{omslaan} (change)). Hence, we created a list of exceptions. If the word was part of a larger expression in the list of exceptions, we removed it from the list of candidate unreported incidents.
We also removed unrelated longer words that contain our keywords (e.g. \emph{weduwe} (widow) contains \emph{duw}, but is unrelated to the word \emph{duw} (push)).
The keywords ``slaat'', ``slaan'', ``geslagen'' and ``sloeg'' are all variants of the verb ``slaan'' (to hit). They are kept separate because the common string ``sl'' is too short and using it as a keyword would lead to too many false positives. Since we already include them in our search, these inflections are required to match full words in the data (not parts of words).
  \begin{table}[tb]
    \centering
\caption{Numbers of admission periods with texts containing potential unreported violence incidents, for various keywords associated with such incidents.}
\label{tab:unreported_incidents_keywords}
\begin{tabular}{l|l|c|c}
  \hline
  Word & English & Exact & Admission periods \\
  \hline
    trap & kick & no & 472 \\
  geslagen & hit (participle) & yes & 415 \\
  slaat & hits (3rd person) & yes & 415 \\
  slaan & hit (infinitive) & yes & 414 \\
  sloeg & hit (past) & yes & 408 \\
  duw & push & no & 343 \\
  schop & kick & no & 299 \\
  bijt & bite & no & 240 \\
\hline
\end{tabular}
\end{table}  

  Table~\ref{tab:unreported_incidents_keywords} shows, for each keyword, how many negative admission periods in our dataset correspond to patients for whom a practitioner note containing that keyword was collected within that same 27-day frame (after exclusion of exceptions). These are candidate admission periods that were labeled as negative but are positive. The total number of such admission periods for all keywords is 1663 (the numbers on the table do not add up to 1663 because some notes contain multiple keywords). 
  This represents 43\% of our 3855 negative admission periods (see Section~\ref{sec:dataset}). 
  
  Upon visual inspection of some of the notes, we found that the large majority of them did not report violence incidents. A lot of them described hitting objects without causing any harm\footnote{Incidentally, this might explain the high correlation between violence incidents and the word \emph{deur} (door) found in~\cite{vincent6}.}; some of them described past events; others talked about hypothetical incidents. However, some cases were very clearly unreported violence incidents.
Part of the challenge in selecting relevant notes is that what constitutes a violence incident is not well defined, as different practitioners have different standards. Therefore, an objective definition of what behavior constitutes violence is needed.

In this paper, we have assumed that the number of unreported violence incidents is small. Researchers and practitioners at UMC Utrecht make use of the violence incident reports frequently, and if the number of unreported incidents were large, they would have encountered problems before. However, further work needs to be done to mitigate this problem carefully.

  \subsection{Feature Importance}
 
\begin{figure}
\centering \includegraphics[width=0.9\textwidth]{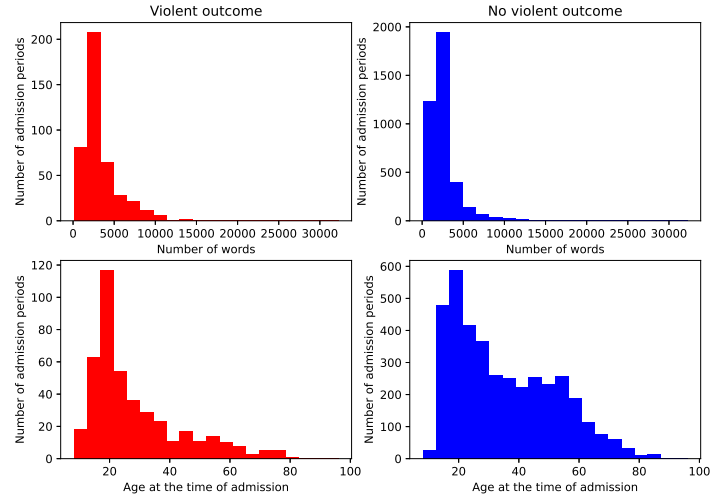}
\caption{Histograms of number of words per period note (top) and age at the beginning of the admission period (bottom) for positive/violent (left/red) and negative/non-violent (right/blue) admission periods.} \label{fig:feat_imp}
\end{figure}

  When using the random forest classifier, we stored the ten most important features according to the best fit in the inner cross-validation loop for each iteration in the outer cross-validation loop. Gathering the most important features together, we then studied both the ten most repeated features and the ten features with the highest total feature importance; these lists were reassuringly similar. The most repeated features were the age at the beginning of the admission period
  and the number of words in the period note.
  The frequency distributions of these variables are shown, for both positive and negative samples, in Figure~\ref{fig:feat_imp}.

As can be seen in the figures, the admission periods with a violent outcome correspond to younger patients on the average, and the average period note with a violent outcome is longer than the average period note with a non-violent outcome.
The fact that only two of the structured variables included in our study resulted in significant discrimination between the positive and negative classes further stresses that novel, sophisticated methods are required, where more careful feature engineering could lead to better discrimination.

\subsection{Limitations}
In this work we used the results from the BVC as a benchmark to the predictions of our machine learning models, because the BVC is currently used for VRA at the University Medical Center (UMC) Utrecht. We found that the performance of some of our models was consistent with that of the BVC, but it should be noted that we computed the AUC for the BVC using violence incidents from the 27 days after the first day of admission.
Therefore, our analysis does not compare with the standard use of the BVC, which is meant to predict violence within 24 hours. Additionally, both datasets seem to be drawn from different distributions, as the class imbalances differ from each other significantly. Efforts to align both datasets were out of the scope of this work.
Furthermore, in our analysis we found indications that the dataset used for the machine learning models has some cases that were not reported as violent, while the associated texts seem to indicate that a violence incident did occur. We do not know the rate of unreported cases, and future work should explore this problem further. 
Lastly, in this work we used BERTje to perform classification on the clinical notes. BERTje is not suited to perform classification on long texts, like clinical notes. Therefore, we experimented with two shortening strategies to fit the requirements of the model: summarization and truncation of texts. Each shortening strategy imposes information loss as not all the original text is used to train the BERTje model.
This loss is likely to contribute to the relatively low performance of the model. Recently, a new BERT-like mode, called SMITH~\cite{yang2020beyond}, has been released that is designed to analyze longer texts. Therefore, an implementation of SMITH for VRA can potentially improve the predictions.

\section{Conclusions}
\label{sec:conclusions}
We applied conventional and deep machine learning methods to the problem of Violence Risk Assessment (VRA), using Dutch-language clinical notes from the psychiatry ward of the University Medical Center (UMC) in Utrecht, the Netherlands.
Our results, reported on Tables~\ref{tab:results} and~\ref{tab:bert_results}, were competitive with a study based on structured variables that obtained AUC = 0.7801~\cite{suchting}, and with the Br{\o}set Violence Checklist (BVC), currently used at the UMC, which gave AUC=0.761$\pm$0.100 when predicting violence incidents within 27 days after a questionnaire was filled out.
These metrics show modest performance, and they indicate that further work is needed to extract all the meaningful information in the clinical notes. 

We have applied a BERT-like model to VRA using clinical notes in Dutch. We found no improvement from using BERTje as opposed to conventional machine learning methods; indeed, the results from BERTje were worse, with AUC$\approx$0.66. Yet we know that our dataset is small and may not be sufficient to fine-tune the large number of parameters present in BERTje. A larger dataset would be highly beneficial but is not trivial to accomplish.

To enlarge the dataset, data from multiple institutions can be aggregated via federated learning, whereby different institutions train the same central model without sharing the data with each other; this is very important, considering privacy restrictions. Additionally, it would be very beneficial to pre-train a `medical-BERTje' model, using Dutch clinical notes from various medical domains.

Finally, a key assumption in our work was that the number of unreported violence incidents was small. The validity of this assumption needs to be scrutinized. Further work is needed to refine the process of selecting unreported incidents from practitioner notes. If the number is small, the datapoints could be removed from the dataset.
This could also inform a re-evaluation of the violence incident reporting in practice. We have seen that practitioners can be very subjective in reporting violence incidents. A unified strategy for incident reporting informed by machine learning could significantly improve data quality.

\section{Acknowledgements}
We acknowledge the COVIDA funding provided by the strategic alliance TU/e, WUR, UU and UMC Utrecht, which funded this study.

\bibliographystyle{cunsrt}
\bibliography{thebibliography}

\end{document}